\documentclass[sn-mathphys]{sn-jnl}

\jyear{2021}%

\theoremstyle{thmstyleone}%
\theoremstyle{thmstyletwo}%

\theoremstyle{thmstylethree}%

\usepackage{graphicx}
\usepackage{subfigure}
\usepackage[justification=centering]{caption}
\usepackage{CJKutf8}
\usepackage{hyperref}
\hypersetup{hidelinks,
	colorlinks=true,
	allcolors=black,
	pdfstartview=Fit,
	breaklinks=true}

\begin{document}
	
	\title[Adaptive Prompt Learning-based Few-Shot Sentiment Analysis]{Adaptive Prompt Learning-based Few-Shot Sentiment Analysis}
	
	
	\author[1]{\fnm{Pengfei} \sur{Zhang}}
	\author[1]{\fnm{Tingting} \sur{Chai}}
	\author*[1]{\fnm{Yongdong} \sur{Xu}}\email{ydxu@hit.edu.cn}
	\equalcont{These authors contributed equally to this work.}
	
	\affil[1]{\orgdiv{school of Computer science and technology}, \orgname{Harbin Institute of Technology at WeiHai}, \orgaddress{\street{2\# WenHuaXi Road}, \city{WeiHai City}, \postcode{264209}, \state{ShanDong Province}, \country{China}}}
	
	\abstract{In the field of natural language processing, sentiment analysis via deep learning has a excellent performance by using large labeled datasets. Meanwhile, labeled data are insufficient in many sentiment analysis, and obtaining these data is time-consuming and laborious. Prompt learning devotes to resolving the data deficiency by reformulating downstream tasks with the help of prompt. In this way, the appropriate prompt is very important for the performance of the model. This paper proposes an adaptive prompting(AP) construction strategy using seq2seq-attention structure to acquire the semantic information of the input sequence. Then dynamically construct adaptive prompt which can not only improve the quality of the prompt, but also can effectively generalize to other fields by pre-trained prompt which is constructed by existing public labeled data. The experimental results on FewCLUE datasets demonstrate that the proposed method AP can effectively construct appropriate adaptive prompt regardless of the quality of hand-crafted prompt and outperform the state-of-the-art baselines.$\footnote{Our implementation is publicly available at \href{https://github.com/simonZPF/AP}{https://github.com/simonZPF/AP}. }$}

	\keywords{Natural language processing, Sentiment analysis, Adaptive prompt learning, Seq2seq-attention}
	
	
	
	\maketitle
	
	\section{Introduction}\label{sec1}
	
	Nowadays, deep learning (DL) has been widely used in image, voice, text and other fields to solve all kinds of problems and get excellent results. At the same time, the model effectiveness depends on large-scale high-quality labeled data which is insufficient. In addition, manual labeled large-scale data is time-consuming and laborious, it is difficult to obtain desirable labeled data to train the model. In order to address data acquisition issue, large-scale unsupervised data and a small amount of supervised data is the first choice for learning, such as semi-supervised learning method. Besides, learning general features from large-scale data, and then adjust them on specific tasks, such as fine-tune pre-trained model and prompt learning. 
	In this work, an adaptive prompt method (AP) by introducing seq2seq-attention structure is proposed to achieve state-of-the-art performances in low resource tasks. In addition, the ability of prompt construction can be further improved by pre-training on the existing labeled datasets in other fields. 
	
	\section{Related work}\label{sec2}

	\subsection{Sentiment analysis}\label{subsec1}
	
Sentiment analysis originates from the analysis of subjectivity in sentences\cite{bib1}. Due to the emergence of a large number of network resources, the research of sentiment analysis has become an active field since 2000\cite{bib2}. Early sentiment analysis mainly focused on building an sentiment dictionary for text classification. It was constructed manually by summarizing words containing sentiment tendencies, and labeling the sentiment polarity and intensity of these words to varying degrees. Therefore, it is necessary to build a high-quality sentiment dictionary\cite{bib3}. Due to the flexibility and non-standard of language, it is difficult to construct a general and efficient rule applicable to all contexts. Machine learning based sentiment analysis mainly relies on NLP researchers or engineers to use their domain knowledge to define and extract significant features from the original data, such as n-gram features, and then use traditional machine learning classifiers such as support vector machine, naive Bayes and maximum entropy for supervised learning\cite{bib4}.Li, G\cite{bib5} builds a model with the prior knowledge of the categorization information in order to extract meaningful features from the unstructured texts by using TF-IDF, short for term frequency-inverse document frequency.

 In recent years, with the development of deep learning theory, neural network has gradually matured in the field of sentiment analysis. Deep neural network can effectively capture the high-level semantic information of text without complex feature engineering, and the expression ability index of the model is times better than that of the shallow model. Among them, convolutional neural network and recurrent neural network are the most widely used\cite{bib6}.Li, D\cite{bib7}proposed the BLSTM and CNN Stacking Architecture (BCSA) to enhance the ability to recognition emotions. Besides, Chen\cite{bib8} propose HUSN which have a novel sentiment classification algorithm that utilizes user’s review habits to enhance hierarchical neural networks.Sadr, H\cite{bib9} proposed model employs recursive neural network due to its tree structure as a substitute of pooling layer in the convolutional network with the aim of capturing long-term dependencies and reducing the loss of local information.

	\subsection{Pre-trained model}\label{subsec2}
	
The purpose of pre-trained language models (PLMs) is to use a large number of texts that have appeared in people's life to train the model, so that the model can learn the probability distribution of each word in these texts, so as to model the model that conforms to these text distributions. 

Traditional PLMs technology aims to to learn word embedding. Because downstream tasks no longer need to use these models, they are usually very low in computational efficiency, such as skip gram\cite{bib10} and glove\cite{bib11}. Although these pre-trained word vectors can capture the semantic meaning of words, they are context independent and can not capture the high-level concepts of text, such as grammar and semantics. Elmo\cite{bib12} proposed a context sensitive text representation method, which constructs the text representation through the deep bidirectional language model, which effectively solves the problem of polysemy. In 2018, Devlin et al\cite{bib13}. proposed BERT (bidirectional encoder representations from transformers) pre-trained language model. The model trains massive corpus through bidirectional transformer encoder and uses masked language model (MLM) to generate in-depth bidirectional language representation. After pre-training, you only need to add an additional output layer for fine-tuning to achieve the performance of state of the art in a variety of downstream tasks. In this process, there is no need to make task specific structural modifications to Bert. 

Bert has opened a new era, and a large number of pre-trained language models have emerged since then. For example, Roberta\cite{bib14} retains the original Bert architecture with longer training time, larger batch, longer sequence and more data. At the same time, delete the prediction of the next sentence and use dynamic masking. Albert\cite{bib15} solves the problems of higher memory consumption and slow Bert training speed. ERNIE\cite{bib16} introduced the knowledge mask strategy, including entity level mask and phrase level mask, to replace the random mask in Bert. In addition,Bert also can be used in other languages. Farahani\cite{bib17} proposed a monolingual Bert for the Persian language which is lighter than the original multilingual Bert model. There is Bert-wwm\cite{bib18} for Chinese which is not only a continuous mask of entity words and phrases, but also a continuous mask of all words that can form Chinese words. 

	\subsection{Prompt Method}\label{subsec3}
	
With the increasing volume of pre-trained language model, the hardware requirements, data requirements and actual cost of fine-tune are also rising. In addition, the rich and diverse downstream tasks also make the design of pre-training and fine-tuning stage cumbersome and complex. Therefore, researchers hope to explore smaller, lighter, more universal and efficient methods. Prompt method is an attempt in this direction which include hand-crafted prompt method and automated prompt. Prompt method is a technology that adds additional text to the input segment in order to better use the knowledge of the pre-trained language model. Schick T\cite{bib19} et al. designed Pattern Exploiting Training(PET), which is a semi-supervised training task. The input example is redefined as the phrase of cloze to help the language model understand the given task. Jiang et al.\cite{bib20} proposed a mining based method, which can automatically find a given set of training input and output templates. This method finds the intermediate word or dependency path between input and output in a large text corpus containing input and output strings, and uses the frequently occurring intermediate word or dependency path as a template.

 Davison et al.\cite{bib21} designed an input (head relation tail) template using LM by studying the tasks related to the knowledge base. Liu x et al.\cite{bib22} proposed a method called P-tuning, abandoned the conventional requirement that ``the template is composed of natural language", used the token never seen in the model to form the template, transformed the construction of the template into a continuous parameter optimization problem, and realized the automatic construction of the template. 
 
In the hand-crafted prompt method, the accuracy of the model depends very much on the quality of the constructed template, and the effects of different templates may vary greatly. For some tasks, it is not so easy to discover an optimal prompt manually. In the automated prompt method, the prompt is constructed by the model and does not rely on manual work. But both of those method can not use the semantic information of input texts in prompt construction process. 

To solve the above problems, this paper proposes a template construction method by introducing seq2seq-attention structure which can dynamically generate matching template vectors and makes full use of the original text information. At the same time, with the idea of pre-trained model, we design a template construction strategy based on pre-training, which can make full use of the public sentiment analysis datasets of high resources fields and apply it to the in other field of low resources.
 
The main contributions of this paper include three aspects:
\begin{enumerate}[1.]
	\item We propose an adaptive prompt method by introducing seq2seq-attention structure. This method has the advantages of both hand-crafted prompt and automated prompt and can make full use of semantic information of input text.
	
	\item The experimental results on the FewCLUE datasets show that the proposed method is effective in the sentiment analysis few-shot task.
	
	\item We proposed to pre-train the adaptive prompt module in high resources tasks, and migrate to or fine-tune in low resource tasks which can effectively play a significant effect in low resource tasks.
\end{enumerate} 

\section{Methodology}\label{sec3}

In this section, we propose an adaptive prompting method based on seq2seq-attention(AP) and introduce its implementation. We introduce seq2seq-attention structure to generate adaptive template from input, and then use the pre-training model to realize sentiment analysis. 

\subsection{Adaptive prompt learning}\label{subsec4}

Our work is based on adaptive prompt learning model, which improve traditional  hand-crafted prompt learning method (HPL). The HPL model includes input layer, hand-crafted prompt, pre-trained language model and output layer. 

Given a pre-trained model \emph{M}, vocabulary \emph{V}, a input sequence 
\textbf{X} of length \emph{n}:\{\emph{x$_{1}$},\emph{x$_{2}$}...\emph{x$_{n}$}\},
verbalizer \textbf{W}:\{\emph{w$_{1}$},\emph{w$_{2}$}...\emph{w$_{l}$}\},
a hand-crafted prompt \textbf{P}:\{\emph{p$_{1}$},\emph{p$_{2}$}...\emph{p$_{i}$},\emph{[MASK]},\emph{p$_{i+1}$}...\emph{p$_{m}$}\},
where value of [MASK] comes form \textbf{W}. Firstly, The sequence input \textbf{X}, and manual prompt \textbf{P} form a template \textbf{t}. Then, the template \textbf{t} will be mapped into e(\textbf{t}):
\{\emph{e(p$_{1}$)},\emph{e(p$_{2}$)}...\emph{e(p$_{i}$)},\emph{e([MASK])},\emph{e(p$_{i+1})$}...\emph{e(p$_{m}$)},\emph{e(x$_{1}$)},\emph{e(x$_{2}$)}...\emph{e(x$_{n}$)}\},by pre-trained  model embedding layer \emph{e}(where each token \emph{p$_{1}$} will be mapped into \emph{e(p$_{1}$)}).After that, the pre-trained model \emph{M} is used to calculate the probability values of [MASK] in \emph{e}([MASK]) to select the best word in \textbf{W} which has the maximum probability. For example, for the sentiment calculation of ``The weather is very good", hand-crafted prompt ``it is [MASK] ", verbalizer \{“good”, “bad”\}, and then the traditional prompt model will construct the template ``It is [MASK], The weather is very good. " Finally, \emph{M} will return the predictive value. As shown in Figure 1(a).  
\begin{figure}[htbp]
	\centering
	\subfigure[hand-crafted prompt method]{
		\begin{minipage}{6cm}
		\centering
		\includegraphics[scale=0.23]{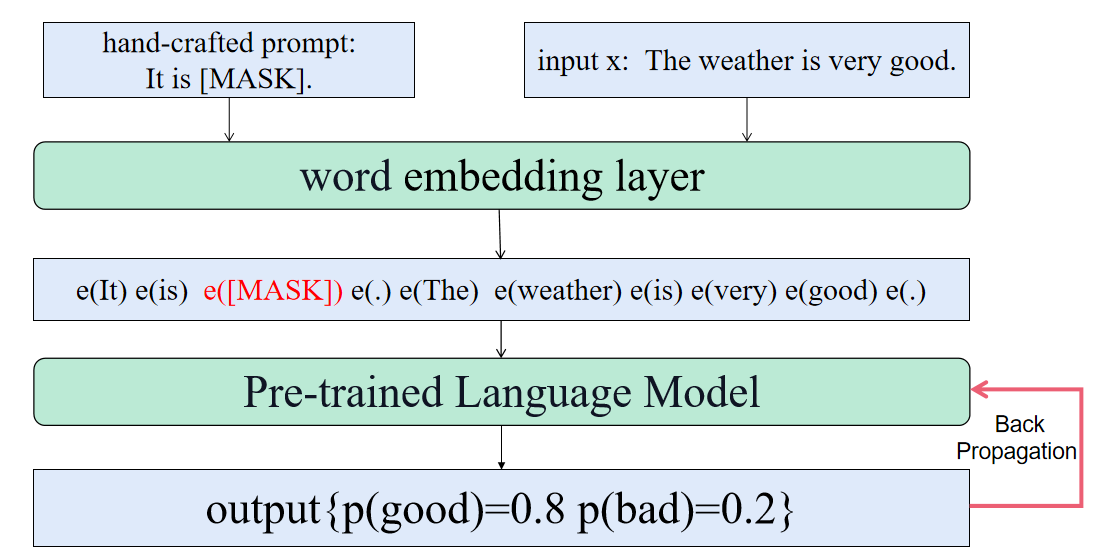}
		\end{minipage}
	}
	\subfigure[adaptive prompt method]{
		\begin{minipage}{5cm}
		\centering
		\includegraphics[scale=0.23]{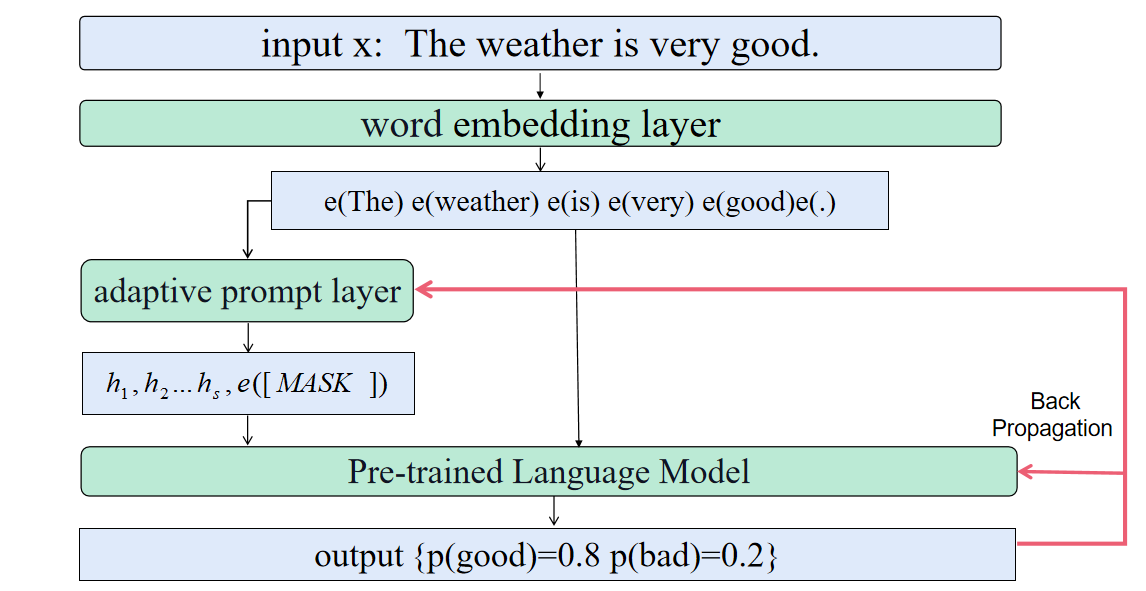} 
		\end{minipage}
	}
	\caption{In (a), the researchers should design prompt by their experience; on the contrary, in (b) the prompt can be automated constructed by adaptive prompt layer.}
\end{figure} 

While the strategy of traditional hand-crafted prompt learning is intuitive and work well in some sentiment tasks, there are also two issues with this approach. 1) It is hard for human to discover optimal prompts in all tasks. 2)Even if in the same task, the optimal prompts of different input sequences are different. Usually, it is hard for algorithms to dynamically found “best” prompt for a special input sequences in task. We consider automatic adaptive prompt design to solve the defects of manual prompt design. As shown in figure 1(b), we use adaptive prompt layer to generate prompt instead of manual craft prompt. In order to strengthen the relationship between the prompt and the input sequence \textbf{X}, we consider to use the text information of the input sequence \textbf{X} to automatically generate the corresponding prompt, that is, generate an adaptive prompt. (in traditional automated method, there is no direct correlation between prompt and input \textbf{X}, we trained the adaptive prompt using the context information of input \textbf{X}). In this work, we use seq2seq-attention structure as adaptive prompt layer to generate adaptive prompt. The seq2seq structure can generate target sequence (prompt sequence) by a specific method according to a source sequence(words embedding vectors of input \textbf{X}) which are suitable for generating prompt sequence. Meanwhile,an attention structure is introduced to better capture the semantic details of input \textbf{X} as shown in Figure 2. 

\begin{figure}[htbp]
	\centering
	\includegraphics[scale=0.4]{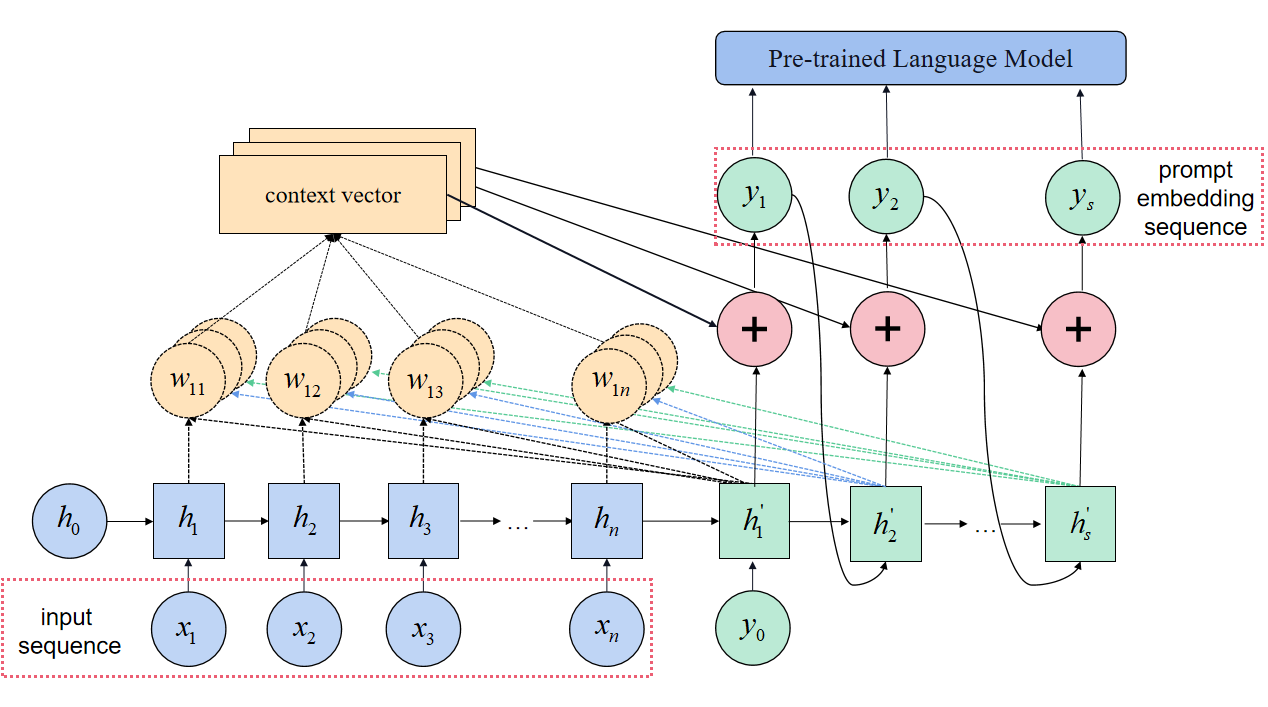}
	\caption{Seq2seq-attention structure, as the adaptive prompt layer of our model. Input is embedded words of \textbf{X}, and output is prompt embedding vector sequence.}
\end{figure} 

In general, seq2seq-attention structure as adaptive prompt layer to generate a vector sequence \textbf{h} with a length of \emph{s} consist the adaptive prompt from input \textbf{X}. And each vector dimension of the sequence is consistent with the output vector dimension of the embedding layer of the pre-trained model. Meanwhile, the adaptive prompt’s embedding vectors are continuous which enables us to find a better continuous prompt beyond the original vocabulary could express\cite{bib22}. In addition, adaptive prompt layer can capture text information of input x by attention structure(the yellow part in the figure 2), which can make the generated prompt fitter with the input text. 
\subsection{Hybrid prompt learning}
Although automated prompt has various advantages in most tasks, such as wide application range, strong generalization ability, stable and balanced performance, the algorithm may falls into local optimal solution in many cases. The hand-crafted prompt has excellent performance in some cases, but this method requires experienced expert participation and is unstable. 
In order to combine the advantages of the two methods, we design a Hybrid prompt composed of a hand-crafted part and automated part as shown in Figure 3. 
\begin{figure}[htbp]
	\centering
	\includegraphics[scale=0.5]{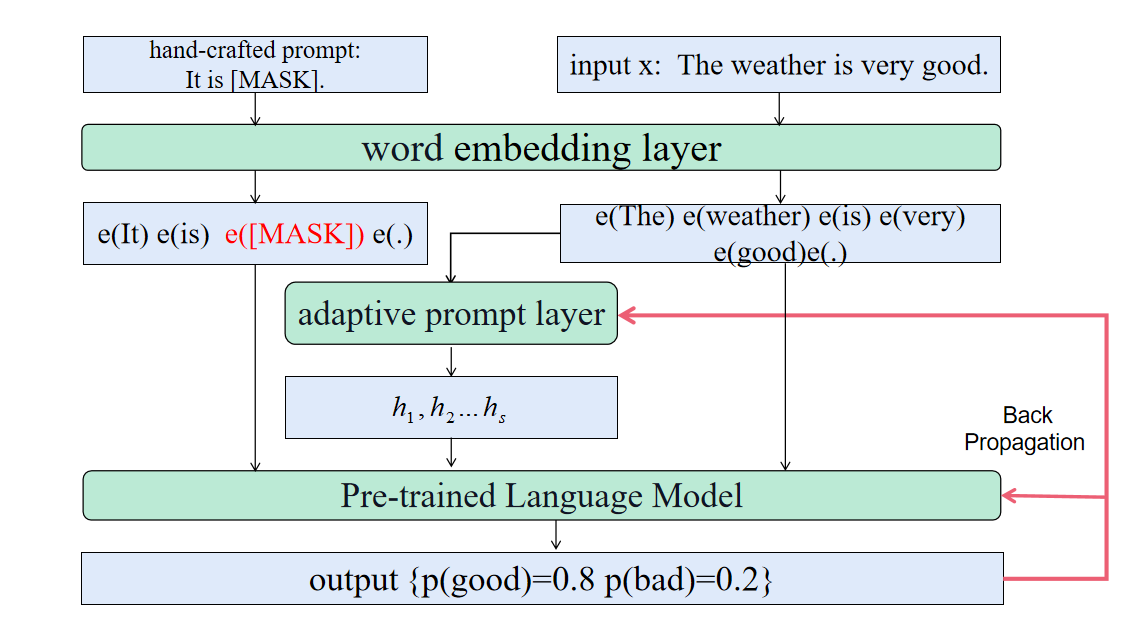}
	\caption{Hybrid prompt model.}
\end{figure} 

The hybrid prompt model combines the word vector generated by hand-crafted prompt and the prompt layer. Through the hybrid prompt embedding layer, the template can be represented a triple $<$ \textbf{X}, \textbf{P}, \textbf{h} $>$=
\{\emph{e(p$_{1}$)},\emph{e(p$_{2}$)}...\emph{e(p$_{i}$)},\emph{e([MASK])},\emph{e(p$_{i+1})$}...\emph{e(p$_{m}$)},
\emph{h$_{1}$},\emph{h$_{2}$}...\emph{h$_{s}$},
\emph{e(x$_{1}$)},\emph{e(x$_{2}$)}...\emph{e(x$_{n}$)}\} 

where \textbf{P} is the hand-crafted prompt, \textbf{h} is the adaptive prompt embedding vector sequence , \textbf{X} is the input text sequence. And the final model prediction result set $y=\{p(i),i\in W\}$  is calculated by pre-trained model. 

Both adaptive prompt and hand-crafted prompt effect the final result \emph{y}. The result showed that(4. 4. 1 for details), the model can learn to adjust the weights of \textbf{P} and \textbf{h} to generate better output results. Therefore, theoretically, the model has the advantages of both hand-crafted prompt and automatic prompt. When a “good” hand-crafted prompt can be found, the model effect can be further improved. Even if a “good” hand-crafted cannot be found, an excellent prompt can be generated by the adaptive prompt part.
\section{Experiments and results}\label{sec4}
\subsection{Database}
We evaluate AP mainly on the FewCLUE public datasets EPRSTMT(E-commerce Product Review Dataset for Sentiment Analysis) task which is labelled as Positive or Negative and collect by ICIP Lab of Beijing Normal University. The datasets used in migration experiment include social media public sentiment datasets(more than 100000 data with emotional labels on Sina Weibo, and about 50000 positive and negative comments respectively), hotel comment data(more than 7000 hotel review data, more than 5000 positive and 2000 negative reviews), user comments data by a take-out platform( 4000 positive and 8000 negative user comments collected by a takeout platform), online shopping data which have 7 categories (books, fruits, shampoo, water heater,milk, clothes and hotels) and more than 60000 comments in total, with about 30000 positive and negative comments respectively. The data in English field include 7000 movie datasets with about 3500 positive and 3500 negative data respectively.
\subsection{Hyper-parameters setting}
In order to fully obtain all the information of the sentence, this paper sets the maximum length of the sentence to twice the length of the digits in the data set. In the experiment, the pre-trained model adopts Roberta-wwm-ext as pre-trained language model. The batch size value is set to 5 and output length of adaptive prompt layer is set to 2 in Chinese and 4 in English, The model adopts Adam optimizer and adopts different learning rates for different optimization methods. 
\subsection{Optimization strategy}
The model consists of two trainable parts, one is the pre-trained model parameters, the other is the seq2seq-attention parameters, that is, the adaptive prompt layer. Based on this, our optimization methods can be divided into two categories: one is to fine-tune all parameters (prompt+LM tuning).In this setting, there are prompt-relevant parameters, which can be fine-tuned together with the all or some of the parameters of the pre-trained models\cite{bib23}.
And the other is to fine-tune only the seq2seq-attention part (fixed LM prompt tuning). In the scenario where additional prompt-relevant parameters are introduced besides parameters of the pre-trained model, fixed-LM prompt tuning updates only the prompts’ parameters using the supervision signal obtained from the downstream training samples, while keeping the entire pre-trained LM unchanged\cite{bib23}. 
\subsection{Results analysis}
\subsubsection{Prompt+LM Tuning}
In this experiment, we use the method of full model parameter adjustment to test the datasets from the FewCLUE in Chinese (32 training sets and about 600 test sets) and the datasets of movie field in English(32 training sets and 600 test sets). In this method, the fine-tuning of pre-traind model plays a leading role in the overall model training, and the seq2seq part plays an auxiliary role. That is, the hand-crafted prompt plays a major role, while the automated template is a supplement and enhancement to the hand-crafted prompt. In this case, the learning rate is 1e-5. The results are shown in Tables 1 and 2. 

\begin{CJK}{UTF8}{gbsn}
\begin{table}[h]
	\begin{center}
			\caption{Accuracy of different methods under different hand-crafted prompt in Chinese datasets}%
			\begin{tabular}{@{}llll@{}}
				\toprule
				Prompt  & Zero-Shot   & HPL  & AP \\
				\midrule
				\_\_开心(happy)    & 75.7\%    & 83.8\%   & \textbf{84.1\%}   \\
				\_\_高兴(glad)  & 70.2\%    & 79.3\%   & \textbf{83.9\%}   \\
				\_\_好(good)   & 64.9\%    & 80.3\%   & \textbf{82.8\%}   \\
 				\_\_行(OK)     & 51.8\% & 78\% & \textbf{82.3\%} \\			
				\botrule
			\end{tabular}
	\end{center}
\end{table}
\end{CJK}
\begin{table}[h]
	\begin{center}
		\caption{Accuracy of different methods under different hand-crafted prompt in English  datasets}%
		\begin{tabular}{@{}llll@{}}
			\toprule
			Prompt  & Zero-Shot   & HPL  & AP \\
			\midrule
			It was \_\_.     & 62.3\%    & 72.7\%   & \textbf{77.8\%}   \\
			Just \_\_! & 60.2\%    & 75.7\%   & \textbf{78.3\%}   \\
			It makes me feel \_\_ that   & 72\%    & 74.5\%   & \textbf{80\%}   \\			
			\botrule
		\end{tabular}
	\end{center}
\end{table}
Among them, the zero-shot method only uses prompt to construct the template, and then predicts through the pre-trained model without fine-tuning the parameters of the pre-trained model. We use the results of zero-shot to judge the quality of hand-crafted prompt. 
In this case, it can be seen that the quality of hand-crafted prompt has a greater impact on HPL method, but less impact on AP method. AP method can also have a higher accuracy when the  hand-crafted prompt is not good. On the other hand, when the hand-crafted prompt is good, AP model can also play a better role than HPL model. Thus, the model can learn to adjust the weights of hand-crafted prompt and adaptive prompt to return better results. 
\subsubsection{Fixed-LM Prompt Tuning}
In order to further test the ability of the seq2seq-attention part of the AP method, we canceled the hand-crafted prompt in this part of the experiment, only used seq2seq-attention to generate the prompt, and frozen the parameters of the pre-trained language model. Therefore, the goal of seq2seq-attention structure is to learn the embedding representation of adaptive prompt in pre-trained language model, makes h behaves like the sequence of real text through the embedding layer. 

We designed experiments on large-scale data (microblog data) and small samples (FewCLUE data). In the experiments on large-scale data, the accuracy of the model is more than 92\%, while in the small sample data, the accuracy of the model is only about 65\%. 

For the good performance of the experiment in the case of large data and the poor performance of small sample data, we consider that in the case of large data, due to the sufficient samples, we can learn the adaptive prompt and its embedding representation in pre-trained model through seq2seq-attention structure. In the case of insufficient samples, seq2seq-attention structure is difficult to learn two parts at the same time, therefore, resulting in over fitting. 

Embedded representation can only be learned under large-scale data. And this  experiment has shown that the model has the ability to learn adaptive prompt with sufficient samples. Therefore, in order to verify that the seq2seq-attention structure can learn the generality of embedded representation of pre-trained model and adaptive prompt, we have done migration experiment. 
\subsubsection{Migration Experiment}
Sentiment analysis includes different fields, such as catering, e-commerce and film. Although there are some differences between these fields, they are generally a classification of sentiment. The reason why the past models can not be used directly across fields is that the words and language structures used for emotional expression in different fields are very different, resulting in different parameters of word vector layer and full connection layer. Therefore, a good automated prompt construction structure should be able to learn the adaptive prompt in the general field and perform well in the unknown field. 

In this experiment, we set up a mixed data experiment. In this experiment, the training set mixes the sentiment analysis data sets of 7 categories (books, fruits, shampoo, water heater, milk, clothes and hotels) of online shopping field, microblog field, takeout field, hotel field and uses the datasets of e-commerce field (FewCLUE) as the test set. In this case, the learning rate is 2e-6, The result of each epoch as shown in figures 5. 
\begin{figure}[htbp]
	\centering
		\begin{minipage}{6cm}
			\centering
			\includegraphics[scale=1]{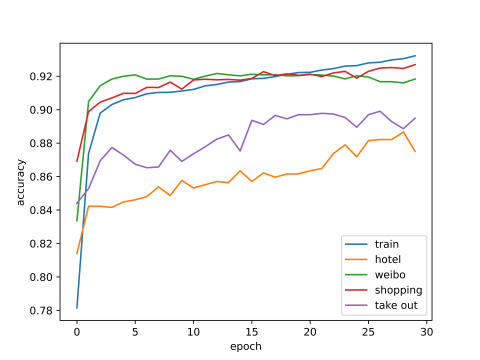}
		\end{minipage}
		\begin{minipage}{5cm}
			\centering
			\includegraphics[scale=1]{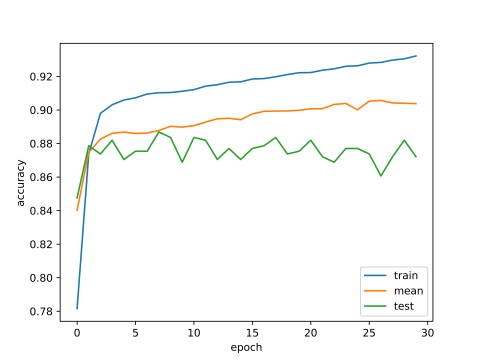} 
		\end{minipage}
	\caption{The accuracy of datasets in different fields in their fields(left),
		and the accuracy of test data(right). }
\end{figure} 

It can be seen that the model can learn to construct general adaptive prompt in the mixed field, and can play a good effect in other fields, with an better accuracy of \textbf{88.7\%} , much higher than the results in 4.1.1 and other method on FewCLUE datasets. As shown in table 3. 
\begin{table}[h]
	\begin{center}
		\caption{Main results of different learning mechanisms on FewCLUE. Values with * are retrieved from Xu et al. \cite{bib24} }%
		\begin{tabular}{l|cccccccc}
		 \toprule
		Method&FineTuning&PET&LM-BFF&P-tuning&EFL&AP\\
		\hline
		Accuracy&65.4\%* &86.7\%* &85.6\%* &88.3\%* &84.9\%* &\textbf{88.7\%}\\
		\botrule
		\end{tabular}
	\end{center}
\end{table}
\subsubsection{Pre-train Experiment}
Pre-training is an application of transfer learning. It uses almost unlimited text to learn the context sensitive representation of each member of the input sentence. It implicitly learns the general grammatical and semantic knowledge, and migrates the knowledge learned from the open domain to the downstream tasks to improve the low resource tasks. We hope to learn the general expression of adaptive prompt from the large-scale sentiment analysis data set through pre-training, so as to better solve the sentiment classification in the case of small samples. 
In this experiment, we pre-train the model in the sentiment analysis data sets of microblog field, takeout field, hotel field and online shopping field. And fine-tuning in the movie field (32 training sets and 600 test sets), in this case, the learning rate is 5e-6. The experimental results are shown in Table 4. 

\begin{table}[h]
	\begin{center}
		\caption{Comparison of experimental results}%
		\begin{tabular}{@{}lll@{}}
			\toprule
			PET  & AP   & Pre-AP  \\
			\midrule
			67.8\%(avg)    & 69.8\%(avg)    & \textbf{78.2\%}     \\
			\botrule		
		\end{tabular}
	\end{center}
\end{table}
The results show that the accuracy of the model is much higher than that of PET and AP models, which means the pre-training method is feasible in AP model. 

\section{Conclusion}\label{sec5}
This paper introduces the method of sentiment analysis, analyzes the shortcomings of prompt learning, and proposed the adaptive prompt model. The advantages of the model can be summarized as follows: 
\begin{enumerate}[1.]
	\item The hand-crafted prompt and automated prompt are combined in the model.
	
	\item Seq2seq-attention structure is introduced to make full use of context information to generate adaptive prompt.
	
	\item The proposed model AP learns to construct a general adaptive prompt by using the sentiment analysis data set in sufficient samples fields.
	
	\item Pre-trained prompt method in the field of sentiment analysis is proposed.
\end{enumerate}

Future research work will be carried out in-depth research from the following aspects to provide directions for further improving the performance of the model: 1) find a better parameter fine-tuning method based on pre-trained prompt;2) This method is extended to other fields of natural language processing, such as text classification, machine reading comprehension.
\bibliography{sn-bibliography}

\end{document}